\documentclass[runningheads]{llncs}

\usepackage{eccv}

\usepackage{eccvabbrv}

\usepackage{graphicx}
\usepackage{booktabs}

\usepackage[accsupp]{axessibility}  % Improves PDF readability for those with disabilities.

\usepackage{hyperref}

\usepackage{orcidlink}

\usepackage{adjustbox}
\usepackage[nohyperlinks, nolist]{acronym} % for acronyms
\usepackage[final]{microtype}
\usepackage{bbm}

\begin{document}

\begin{acronym}

\acro{sae}[SAE]{Sparse Autoencoder}
\acro{llm}[LLM]{Large Language Model}
\acro{wsi}[WSI]{Whole Slide Image}
\acro{mil}[MIL]{Multiple Instance Learning}
\acro{tma}[TMA]{Tissue Microarray}
\acro{vfm}[VFM]{Vision Foundation Model}
\acro{vit}[ViT]{Vision Transformer}
\acro{mse}[MSE]{Mean Squared Error}
\acro{dori}[DoRI]{Domain Robustness Index}
\acro{expose}[EXPOSE]{\textbf{Ex}plainable \textbf{P}robing of Cross-D\textbf{o}main \textbf{S}parse \textbf{E}mbeddings}
\acro{uke}[UKE]{University Medical Center Hamburg-Eppendorf}
\acro{rp}[RP]{Radical Prostatectomy}
\acro{ood}[OoD]{out-of-domain}
\acro{id}[ID]{in-domain}
\acro{auroc5}[AUROC5]{AUROC for 5-year relapse prediction}
\acro{auroc}[AUROC]{Area under the Receiver Operating Characteristic Curve}

% \acro{first}[UKE.first]{UKE.first}
% \acro{scanner}[UKE.scanner]{UKE.scanner}
% \acro{spot}[UKE.spot]{UKE.spot}
% \acro{thin}[UKE.thin]{UKE.thin}
% \acro{thick}[UKE.thick]{UKE.thick}
% \acro{long}[UKE.long]{UKE.long}
\acro{first}[ScanA]{ScanA}
\acro{scanner}[ScanB]{ScanB}
\acro{spot}[ScanA.spot]{ScanA.spot}
\acro{thin}[ScanA.thin]{ScanA.thin}
\acro{thick}[ScanA.thick]{ScanA.thick}
\acro{long}[ScanA.long]{ScanA.long}

\end{acronym}

\title{EXPOSE: Explainable and Domain-Robust Embeddings from Pathology Vision Foundation Models using Sparse Autoencoders}

\titlerunning{EXPOSE}

\author{Anja Witte\inst{1}\orcidlink{0009-0007-8519-7314} \and
Maximilian Lennartz\inst{2}\orcidlink{0000-0002-1572-8200} \and
Jan Baumbach\inst{3}\orcidlink{0000-0002-0282-0462} \and
Guido Sauter\inst{2}\orcidlink{0000-0002-9024-4978} \and
Stefan Bonn\inst{1,4}\orcidlink{0000-0003-4366-5662} \and
Patrick Fuhlert*\inst{1}\orcidlink{0000-0001-8480-3705} \and
Marina Zimmermann*\inst{1,5}\orcidlink{0000-0002-2666-5997}}

\authorrunning{A. Witte et al.}

\institute{Institute of Medical Systems Bioinformatics, Center for Biomedical AI (bAIome), Center for Molecular Neurobiology Hamburg (ZMNH), Hamburg Center for Translational Immunology (HCTI), University Medical Center Hamburg-Eppendorf, Hamburg, Germany \and
Institute of Pathology, University Medical Center Hamburg-Eppendorf, Hamburg, Germany \and
Institute of Computational Systems Biology, University of Hamburg, Hamburg, Germany \and
German Center for Child and Adolescent Health (DZKJ), Hamburg, Germany \and
III. Department of Medicine, University Medical Center Hamburg-Eppendorf, Hamburg, Germany\\
* shared last author \\
\email{marina.zimmermann@zmnh.uni-hamburg.de}}

\maketitle

\begin{abstract}
\acp{vfm} are widely used in computational pathology but remain sensitive to domain shifts arising from variations in staining, tissue preparation, and scanner hardware. A key limitation is that \ac{vfm} embeddings entangle biological with domain-specific information, hindering cross-domain generalization. We propose \ac{expose}, a framework that uses \acp{sae} as an explainable bottleneck to identify and suppress domain-specific components in \ac{vfm} embeddings. We train a sparse representation of \ac{vfm} features, use a linear classifier to identify domain-specific latent dimensions, and mask these features prior to downstream relapse prediction without retraining the backbone model. Experiments on a large prostate cancer dataset with multiple acquisition domains show that \ac{sae} features capture both domain- and task-specific information, which are partially disentangled in the latent space. Removing domain-specific features improves cross-domain performance and increases embedding robustness as measured by the \ac{dori}. Code is available at \url{https://github.com/imsb-uke/expose}.
  
\end{abstract}

\acresetall

\acused{first}
\acused{scanner}
\acused{spot}
\acused{thin}
\acused{thick}
\acused{long}
\acused{expose}

\section{Introduction}\label{sec:intro}

\acp{vfm} have become the de-facto standard for computational pathology, providing powerful transferable representations that achieve strong performance across a wide range of downstream tasks \cite{bilal2025,Li2025}. Their success is driven by large-scale models that are pretrained on massive and diverse datasets, which enable effective feature reuse in data-scarce medical settings \cite{chen2024uni,zimmermann_virchow2_2024,hoptimus0,Nechaev2024}. Despite this progress, \acp{vfm} remain sensitive to domain shifts -- a key challenge in computational pathology \cite{Stacke2021}. Such domain shifts arise from variations in tissue preparation, staining protocols, and scanner hardware and often negatively impact machine learning models \cite{Howard2021}. Consequently, the performance of \acp{vfm} can degrade substantially when deployed in external domains \cite{Kmen2026,Carloni2025, Witte2026}.

One key limitation underlying this behaviour is that \acp{vfm} often encode biologically relevant information together with domain-specific factors in feature representations~\cite{Kmen2026}. While biological signals are essential for diagnostic tasks, domain-specific features can limit model generalization and reduce reliability. Disentangling these factors in high-dimensional representation spaces is challenging because learned features lack semantic meaning, making it difficult to identify which dimensions encode biological or domain-specific information.

\acp{sae} have recently emerged as a promising tool for transforming dense foundation model embeddings into sparse and more explainable representations \cite{Cunningham2023,gatedsae}. By learning an overcomplete and sparse latent basis, \acp{sae} have been shown to recover features that often align with semantically meaningful concepts \cite{bricken2023,Gujral2025}. This provides a mechanism for analysing the internal structure of foundation model representations, enabling analysis and selection of individual latent features \cite{Cunningham2023,pach2026sparse}.

Recent studies in computer vision and computational pathology have shown that \ac{sae} features can capture biologically meaningful structures that are consistent across datasets, suggesting that sparse representations may separate stable semantic information from dataset-specific variation \cite{dasdelen2026,le2025,Sun2025}. At the same time, sparse features have been shown to be amenable to post-hoc manipulation, as individual latent dimensions can be selectively modified or suppressed \cite{Sun2025,Kim2026}. However, it remains an open question whether domain-specific information is distinctly represented within the sparse latent space, and whether such information can be explicitly identified and removed to improve robustness.

\begin{figure}[tb]
    \begin{centering}
    \includegraphics[width=\linewidth]{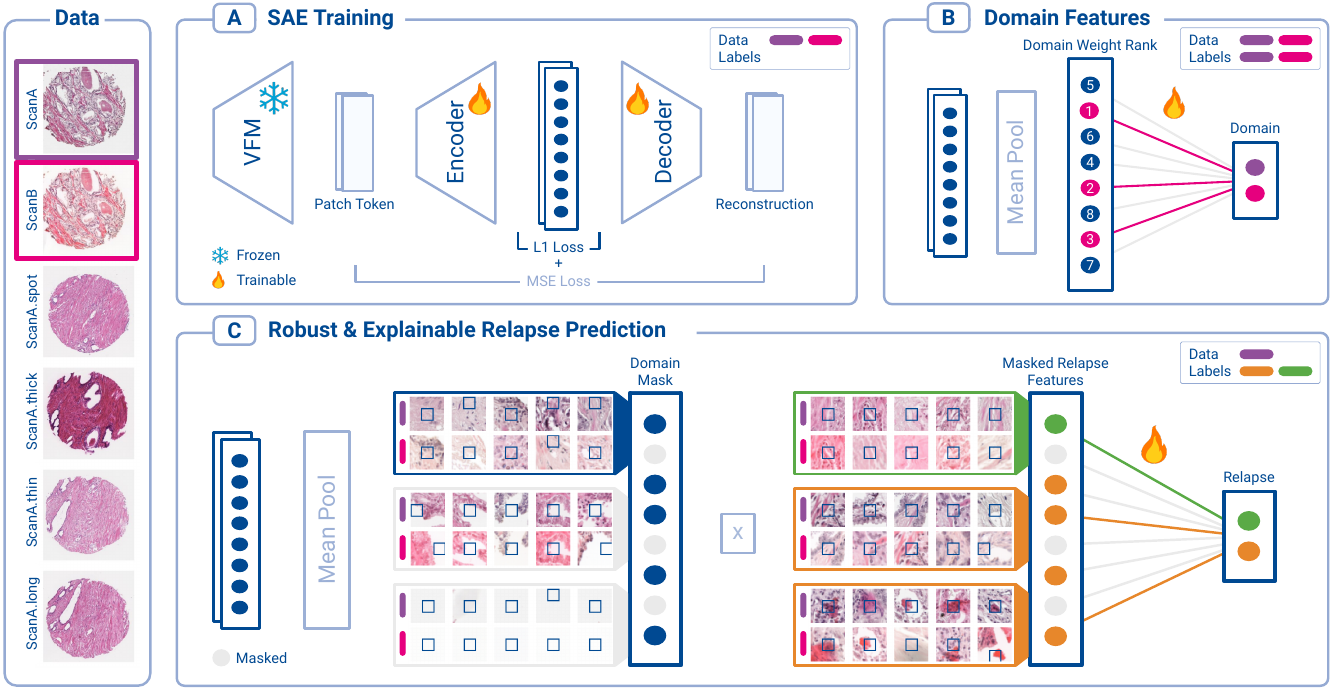}
    \par\end{centering}
    \caption{Overview of \ac{expose}. (A) An \ac{sae} is trained in an unsupervised manner on \ac{vfm} patch embeddings of two domains to learn cross-domain and disentangled representations. (B) The \ac{sae} encoded features are mean-pooled and a linear classifier is trained to predict the domain of these slide-level representations. Feature dimensions are ranked according to their weights in the domain classifier. The features with the top-$k$ highest weights are defined as domain-specific features. (C) The mean-pooled features identified as domain-specific are set to zero. A linear classifier is trained to predict the binary downstream task of relapse prediction. The most relevant features for both domain and relapse classes are linked to visual features. \label{fig:overview}}
\end{figure}

In this work, we address this question with \acs{expose}, a framework for  \acl{expose} (see \cref{fig:overview}). \ac{expose} transforms \ac{vfm} embeddings into a sparse latent space, automatically identifies features associated with domain-specific information, and masks these features before training the downstream model on the resulting filtered representations. In contrast to prior work that uses sparse representations primarily for explainability or semantic feature manipulation, \ac{expose} uses \ac{sae} features as an explainable bottleneck for removing domain-specific information to improve robustness under domain shift.
Our contributions are summarized as follows:
\begin{itemize}
    \item We propose \ac{expose}, a framework that leverages \acp{sae} to obtain explainable feature representations of \ac{vfm} embeddings and enables domain-aware feature masking.
    
    \item We introduce an automatic mechanism for identifying domain-specific latent features in the \ac{sae} space and show that the mechanism relates to domain-specific feature activation patterns. 
    
    \item We demonstrate that masking domain-specific features improves cross-domain robustness and provides evidence that biological and domain-specific information are partially separated in the learned sparse representation. The optimal masking level depends on whether data is drawn from the training domain, \ac{id}, or a shifted target domain, \ac{ood}.

    \item We further show that progressively removing domain features increases biological clustering structure in the latent space, indicating improved separation of semantic and domain-specific information.
\end{itemize}

\section{Related Work}\label{sec:related-work}
This section reviews related work on \acp{sae} for explainable representation learning and domain robustness in computational pathology. We then discuss recent attempts to improve robustness through sparse feature representations and position our approach among existing methods.

% SAE - LLM
\subsubsection{Sparse Autoencoders for Explainability.}
\ac{sae} architectures have been increasingly applied in transformer-based foundation models to facilitate analysis and explanation of model representations by learning sparse feature representations, \eg in \acp{llm}~\cite{bricken2023}. As sparsity encourages neurons with distinct semantics, a property often referred to as monosemanticity, sparse features have been observed to activate consistently for specific semantic concepts \cite{ksparse, huben2024}. Empirical evidence suggests that the overall latent space tends to be hierarchically clustered according to semantic concepts \cite{zhang2025}. The monosemanticity of the \ac{sae} embedding space further allows model behaviour steering, the manipulation of internal model representations at inference, through selecting features that correspond to the concept of interest and modifying its activations \cite{He2025,Cho2025}. These works demonstrate that sparse features are explainable and can serve as targets for post-hoc interventions.

Moreover, \acp{sae} are increasingly applied in the field of computer vision to enhance explainability through monosemantic features \cite{pach2026sparse,le2025,Olson2025}. Beyond explainability, few studies have explored adapting the \ac{sae} embeddings for downstream vision tasks. Examples are visual steering through manipulating a connected text encoder \cite{chatzoudis2025,joseph2025} and causal interventions \cite{stevens2025}.

% VFM and robustness, Computational Pathology
\subsubsection{Domain Shift and Robustness in Computational Pathology.}
In computational pathology, \acp{vfm} have become a prominent approach for AI-guided image analysis \cite{bilal2025, xiong2025}. These large-scale models, \eg \cite{chen2024uni, zimmermann_virchow2_2024, Shao2025}, are usually trained on extensive and diverse datasets during pre-training, leading to higher downstream task performance while aiming to achieve more robust representations. Robustness is particularly critical in pathology as clinical workflows for tissue preparation and digitization introduce biologically irrelevant visual changes which can affect model performance \cite{Stacke2021,guan2022}.  
% robustness
Nevertheless, several studies have shown the limited domain robustness of \acp{vfm} in terms of performance \cite{Campanella2025,Dietrich2021,Fuhlert2026}. Biological- and domain-specific signals are entangled in the learned features, causing the model to potentially rely on non-diagnostic information \cite{Kmen2026}. Accordingly, there is a need for methods to increase robustness of \ac{vfm} representations against domain shifts. While prior work improves \ac{vfm} cross-domain performance through ensembling \cite{Bareja2025}, distillation \cite{Filiot2025}, or embedding alignment \cite{Carloni2025}, these approaches do not explicitly identify domain-specific feature dimensions.

\subsubsection{Increasing Domain Robustness through Sparse Embeddings.}
Recent studies have shown that \acp{sae} can recover partially explainable and monosemantic features from \acp{vfm} activations, which can correspond to medically meaningful concepts in sparse latent spaces \cite{dasdelen2026,le2025}. Such features can be validated across datasets by experts, suggesting that sparse representations may capture recurring biological factors beyond dataset-specific variation.

In medical imaging, manual or supervised interventions in feature space have been used to reduce reliance on spurious artifacts. For example, \emph{ProtoMIL} \cite{Sun2025} introduces expert-driven interventions by identifying visually irrelevant artifacts (\eg, ink or staining artifacts) and masking their corresponding activations, enabling the downstream model to focus on more biologically relevant cues. This line of work raises an important question: can domain-specific information also be localized to individual sparse features in a similar fashion?

Existing \ac{sae}-based approaches in computational pathology primarily use sparse features for explainable downstream prediction, where the identification of relevant factors or the design of feature interventions often depends on manual expertise \cite{le2025,Olson2025}. Consequently, it remains unclear whether domain-specific shortcuts are automatically encoded as identifiable sparse features, and whether suppressing those features improves domain robustness when learning a task model under a constrained domain training setting. 

In this work, we study this hypothesis by learning a linear domain classifier over the sparse latent space and using its weights to derive a feature-level domain importance. We then mask domain-specific sparse features while training a linear relapse classifier on a single domain, and evaluate the cross-domain robustness.

\section{Methods}\label{sec:methods}

Our proposed \ac{expose} framework (visualized in \cref{fig:overview}), that transforms \ac{vfm} embeddings into sparse representations and masks domain-specific features, consists of three steps and is trained on our multi-domain dataset that is described in \cref{sec:methods-dataset}. First, we train an \ac{sae} to map \ac{vfm} patch embeddings into a sparse representation by optimizing a reconstruction objective with $\ell_1$ sparsity regularization (\cref{fig:overview} (A), see \cref{sec:methods-sae}). We then reuse the sparse features as the representation for the two subsequent steps (B) and (C). In (B), we train a binary domain classifier on top of the sparse activations and use its learned linear domain weights to identify the corresponding sparse features (see \cref{sec:methods-domain-classifier}). In (C), we use the same setup to train a linear 5-year relapse classifier on a dataset with a single domain. Before training, we mask the sparse feature dimensions identified as domain-specific by the domain classifier by setting their activations to zero. The relapse classifier is therefore trained on the remaining domain-independent sparse representation, encouraging it to rely on features associated with biological rather than domain-specific variation. Finally, for explainability, we analyse the sparse features and visualize high activated image regions to understand which features drive the model's reasoning (see \cref{sec:methods-relapse-classifier}). 

\subsection{Dataset}\label{sec:methods-dataset}
Our internal prostate cancer dataset contains 69,251 \ac{tma} spot images and patient-specific metadata from 17,700 patients that underwent \ac{rp}. The dataset was collected between 1992 and 2014. For the purpose of this study, we filtered the dataset with respect to image quality and patient-related information. We included only patients with documented follow-up for at least five years or experiencing relapse within five years after \ac{rp}. Relapse is defined as biochemical recurrence, additional unplanned therapy, metastasis, and prostate cancer-related death.

The final dataset contains 24,588 \ac{tma} spot images that are divided into six subdatasets according to different tissue preparation and digitization protocols. Sample images are shown in \cref{fig:overview}. The default subdataset is \ac{first} as it follows our standard acquisition protocol. The other five subdatasets \ac{scanner}, \ac{spot}, \ac{thin}, \ac{thick} and \ac{long} emulate different domain shifts resulting from differences in the tissue preparation and digitization protocol. Detailed information about the differences between the subdatasets can be found in the appendix (\cref{sec:apx-data,tab:apx:dataset-info}). As \ac{first} and \ac{scanner} are used for the training of the domain and the relapse classifier, they are considered as \ac{id}, whereas the remaining four subdatasets represent unseen domains and are therefore \ac{ood}.

\subsection{Sparse Autoencoder Objective}\label{sec:methods-sae}
The goal of the ReLU-based \ac{sae} architecture \cite{ng2011sparseautoencoder} is to learn a sparse representation $h \in \mathbb{R}^{d_{\mathrm{hid}}}$ for each dense input $z$ through embedding reconstruction while $d_{\mathrm{in}} < d_{\mathrm{hid}}$. We chose an expansion factor of $5$ and obtain $d_{\mathrm{hid}} = 5 \cdot 768 = 3840$ sparse dimensions. The \ac{sae} consists of an encoder $E$ and a decoder $D$ that are linear transformations with weight matrices $W_e$ and $W_d$ and biases $b_e$ and $b_d$. The transformations are defined as
\begin{equation}
  h = E(z) = \operatorname{ReLU}(W_e (z - b_{d}) + b_e), \qquad \hat{z} = D(h) = W_d h + b_d,
  \label{eq:sae}
\end{equation}
where ReLU activation ensures non-negative values in the sparse representation $h \in \mathbb{R}_+^{d_{\mathrm{hid}}}$. We train the \ac{sae} to minimize the reconstruction loss of the dense embedding with \ac{mse} while encouraging sparsity in $h$ via $\ell_1$ regularization as
\begin{equation}
L_{\mathrm{SAE}}(z)=\|\hat{z}-z\|_2^2+\lambda\|h\|_1.
\label{eq:sae-loss}
\end{equation}

After training, we freeze the \ac{sae} encoder and use the mean-pooled sparse representation over all patch embeddings of a \ac{tma} spot for the subsequent steps.

\subsection{Identifying Domain Features}\label{sec:methods-domain-classifier}

Given the frozen sparse latent representation $h$ obtained from the \ac{sae}, we train a binary domain classifier to predict whether an embedding originates from the domain \ac{first} or \ac{scanner} by utilizing registered, pairwise images from those domains.

We hypothesize that highly weighted features in the domain classifier correspond to sparse dimensions that are most relevant for domain separation and are associated with domain shift effects that impact downstream performance. Consequently, masking these features should reduce domain-specific bias and improve robustness. Since $h_i > 0 \quad \forall i \in \{1,\ldots,d_{\mathrm{hid}}\}$ due to ReLU activation, the sign and magnitude of the classifier weights provide an interpretable signal for domain association: features with high positive weights are most predictive of \ac{scanner}, whereas features with low negative weights are most predictive of \ac{first}.

Therefore, we rank all sparse features using one of three strategies $s$: highest absolute weight, highest positive weight, or lowest negative weight. According to the strategy $s$, we then select the top-$k$ sparse features and construct a domain mask \mbox{$\mathcal{M} \in \{0,1\}^{d_{\mathrm{hid}}}$} that masking these features.

Further, this mask is utilized for 5-year relapse classification (see \cref{sec:methods-relapse-classifier}) to remove the sparse features that are most responsible for distinguishing the two training domains, thereby explicitly targeting domain-shift-inducing features and encouraging the classifier to rely on domain-invariant evidence.

\subsection{Robust and Explainable Relapse Prediction}\label{sec:methods-relapse-classifier}
Using the sparse latent representation $h$ from the \ac{sae} and the binary domain mask $\mathcal{M}$, we train a linear classifier with sigmoid activation for 5-year relapse prediction using the masked sparse features. For each sample, the masked sparse representation is obtained as $h' = h \odot \mathcal{M}$, where $\odot$ denotes element-wise multiplication. This setup prevents the model from relying on sparse latent features that were identified as most informative for domain discrimination during Step (B), forcing it to use the remaining, less domain-specific sparse features. In contrast to the previous \ac{sae} and domain classifier, the relapse classifier is trained using binary 5-year relapse labels.

\subsubsection{Explainability.} 
We provide feature-level explanations by reporting the sparse features selected by the linear classifiers. For both the domain and relapse classifiers, we identify the sparse feature dimensions with the top negative and top positive classifier weights, corresponding to \ac{first}/\ac{scanner} and no-relapse/relapse predictions, respectively. For each selected sparse feature, representative image examples for domain classification are obtained by retrieving the \ac{tma} spots with the highest mean activation of the respective feature across all patches. For relapse classification, we interpret the selected features at the patch level by visualizing patches that maximally activate the respective feature. This yields an explicit, sparse, and explainable set of feature dimensions associated with domain- and relapse-specific evidence.

\subsubsection{Domain Robustness Index.}\label{sec:dori}
As an individual measure for domain robustness in the embedding space, we use the \ac{dori}, a normalized version of the Robustness Index of \cite{de2025,Kmen2026}. \ac{dori} compares the distribution of the biological target labels to the domain information in the embedding space of a foundation model used for the downstream analysis task.

More formally, consider the $k$ nearest neighbours according to a distance metric (we use cosine similarity) around all $N$ embeddings with $y$ distinct available target labels and $d$ distinct available domains. The mean number of neighbours $\bar{y}$ and $\bar{d}$ with the same label- and domain-specific information, respectively, is calculated as

\begin{equation}
\bar{y}=\frac{1}{Nk}\sum_{i=1}^{N}\sum_{j=1}^{k}\mathbbm{1}(y_i=y_j),
\qquad
\bar{d}=\frac{1}{Nk}\sum_{i=1}^{N}\sum_{j=1}^{k}\mathbbm{1}(d_i=d_j).
\label{eq:sum_same_label}
\end{equation}

Comparing $\bar{y}$ and $\bar{d}$ measures their respective influence on the embedding where $\mathbbm{1}(\cdot)$ is the indicator function yielding $1$ if its argument is true. In contrast to \cite{de2025}, we normalize these quantities since they depend on $y$ and $d$. Let a uniform embedding yield expected values $\mathbbm{E}[y]=y^{-1}$ and $\mathbbm{E}[d]=d^{-1}$. The normalized \ac{dori} is therefore defined as 
\begin{equation}
    \mathrm{\ac{dori}}(\mathbf{y},\mathbf{d})
    =
    \frac{\bar{y}-\mathbbm{E}[y]}{\mathbbm{E}[y]}
    -
    \frac{\bar{d}-\mathbbm{E}[d]}{\mathbbm{E}[d]}.
    \label{eq:dori}
\end{equation}

Here, $\mathbf{y}\in\{1,\ldots,y\}^{N}$ and $\mathbf{d}\in\{1,\ldots,d\}^{N}$ denote the label and domain assignments for all samples. The resulting \ac{dori} is bounded to $[-1,1]$, where positive values indicate that label information has a stronger influence on the embedding than domain information, whereas negative values indicate the opposite.

\section{Experimental Setup}\label{sec:exp-setup}

\subsubsection{Datasets and Splits.}
We evaluate \ac{expose} on the internal prostate cancer dataset described in \cref{sec:methods-dataset}. The \ac{first} and \ac{scanner} subdatasets, each containing 8,141 registered samples, are used for training the \ac{sae}, and the domain classifier. The relapse classifier is trained on \ac{first}. The two subdatasets are split into training (70\%), validation (15\%), and test (15\%) with patient-level stratification based on the 5-year relapse label. The remaining four subdatasets (\ac{spot}, \ac{thin}, \ac{thick}, and \ac{long}) are not used during training and serve exclusively as additional evaluation domains. To avoid leakage, their validation and test splits follow the same patient assignment as the corresponding \ac{first} split. Consequently, \ac{first} and \ac{scanner} are considered \ac{id} domains, whereas \ac{spot}, \ac{thin}, \ac{thick}, and \ac{long} represent unseen \ac{ood} domains.

The \ac{sae} is trained and validated using the training and validation splits of both \ac{first} and \ac{scanner} to learn sparse representations that capture biological structure while disentangling domain-specific variation. The domain classifier is trained on the same two subdatasets and splits, whereas the relapse classifier is trained exclusively on the training and validation splits of \ac{first}. Hyperparameter tuning for all three tasks is performed using the validation split.

\subsubsection{Input Preprocessing.}
All \ac{tma} images are preprocessed by extracting a centred crop of size \(2048 \times 2048\) pixels at \(40\times\) magnification to reduce background regions. The cropped images are downsampled to \(5\times\) magnification. Patch embeddings \(z \in \mathbb{R}^{d_{\mathrm{in}}}\) with \(d_{\mathrm{in}} = 768\) are extracted using H0-mini~\cite{Filiot2025}, a distilled version of the pathology \ac{vfm} H-Optimus-0~\cite{hoptimus0}, and serve as input to the \ac{sae}. Image preprocessing follows the protocol proposed by the authors of H0-mini \cite{Filiot2025}.

\subsubsection{Training Details.}
The \ac{sae} is trained with the AdamW optimizer \cite{adamw} using a learning rate of \(1.5 \times 10^{-4}\), an \(\ell_1\) regularization weight of \(\lambda = 10^{-5}\), a batch size of 128, and early stopping with a patience of 10 epochs (maximum 50 epochs). The learning rate is selected from the range \([10^{-5}, 10^{-2}]\), while the \(\ell_1\) regularization weight is chosen from \(\{10^{-2}, 10^{-3}, 10^{-4}, 10^{-5}, 10^{-6}\}\) using Bayesian search.

The domain and relapse classifiers are optimized using the AdamW optimizer~\cite{adamw} with a binary cross-entropy loss, a batch size of 128, and early stopping with a patience of 10 epochs. The domain classifier uses a learning rate of \(10^{-4}\), whereas the relapse classifier is trained with a learning rate of \(8.5 \times 10^{-3}\), which is tuned over the same search range \([10^{-5}, 10^{-2}]\) as the \ac{sae}. Before training the relapse classifier, the top-\(k\) domain-specific sparse features identified by the domain classifier are masked by setting their activations to zero. We evaluate \(k \in \{2,4,8,16,32,64,128,256,512,1024,2048,3000,3500\}\), covering a range from individual feature masking to suppressing most of the sparse representation dimensions (up to 3500 of 3840 dimensions).

For all models, early stopping is based on validation loss. Unless stated otherwise, results are reported as mean $\pm$ standard deviation over five runs with different random seeds.

\subsubsection{Evaluation Protocol.}
The domain classifier is evaluated using the \ac{auroc} on the held-out test sets of the \ac{first} and \ac{scanner} domains. The relapse classifier is trained exclusively on the \ac{first} domain and evaluated on all six subdatasets to assess both \ac{id} and cross-domain generalization. We compare our approach against a base model that uses frozen \ac{vfm} embeddings and a linear classifier to predict relapse. Performance is measured using \ac{auroc5}. In addition to predictive performance, we report \ac{dori} as an embedding-based measure of domain robustness that is computed using cosine distance with \(k=20\).
\section{Results}\label{sec:experiments}
We evaluate whether \ac{sae} features can help identifying features that carry domain-specific information and how this relates to relapse prediction. First, we identify domain-specific feature dimensions using a linear classifier over \ac{sae} activations and analyse different selection strategies (\cref{sec:results-domain-features}). Second, we compare domain-specific features with relapse-specific features to assess whether both tasks rely on shared or distinct representations (\cref{sec:results-domain-vs-relapse}). Finally, we investigate whether masking domain-specific features improves cross-domain robustness and representation quality (\cref{sec:results-robustness-impact}).

\subsection{Identifying Domain-Specific SAE Features}\label{sec:results-domain-features}

% A linear layer was trained on \ac{first} and \ac{scanner} sparse features to predict the respective domain. The experiment was repeated with five different random seeds. A first observation is that the learned feature weights were highly consistent across runs, with a Spearman rank correlation of $\rho_s=0.975$  ($p < 10^{-4}$), indicating that the importance assigned to individual features for domain classification was nearly identical across training runs (see \cref{apx:???}.

We first evaluate our three strategies for identifying domain-specific \ac{sae} features. Since the domain classifier is linear, sparse features can be ranked according to the sign and magnitude of their classifier weights. Positive and negative weights contribute to predicting \ac{scanner} and \ac{first}, respectively. We compare selecting features associated with \ac{scanner} (top positive weights), \ac{first} (top negative weights), and both domains (top absolute weights). Additionally, we compare all strategies against using the complete \ac{sae} representation.

\begin{figure}[tb]
    \centering
    \includegraphics[width=\linewidth]{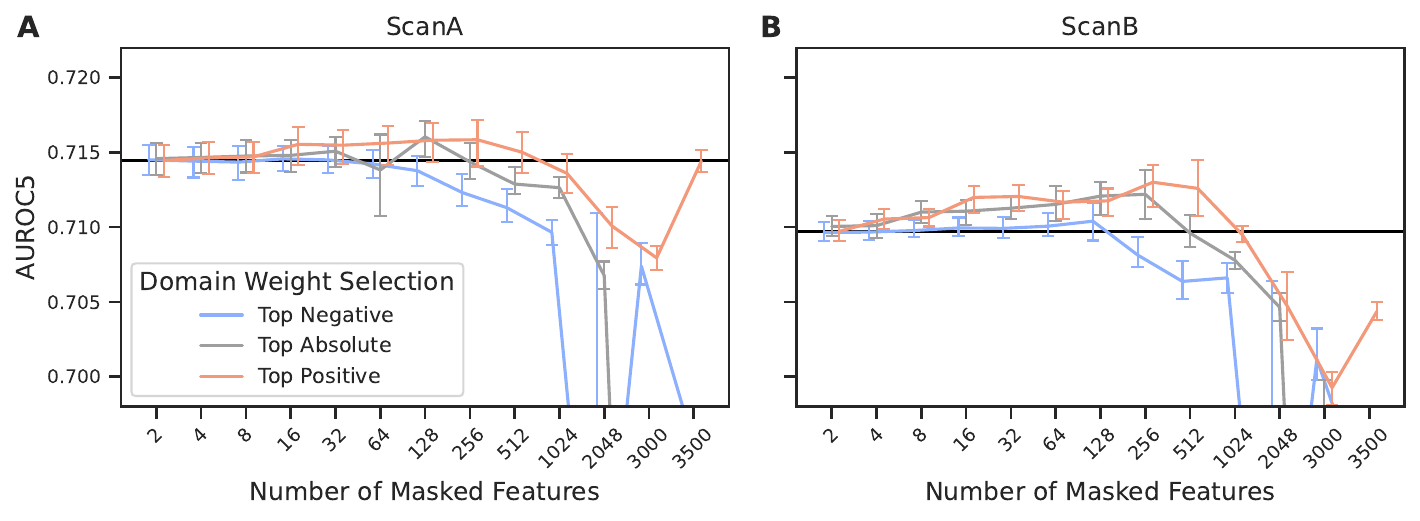}
    \caption{Evaluation of relapse prediction while masking the top-$k$ domain-specific features chosen with different selection strategies based on the linear weights for domain classification. Top negative and top positive weights can be defined as relevant for \ac{first} and \ac{scanner} classification, respectively. Features that are relevant for both domains in the domain classifier are chosen by selecting the features with the top absolute domain weights. Validation results are shown for the evaluation of \ac{first} (A) and \ac{scanner} (B) dataset.}
    \label{fig:topk_selection_strategies}
\end{figure}

\cref{fig:topk_selection_strategies} shows 5-year relapse prediction performance while progressively masking the top-$k$ selected features according to each strategy. For small values of $k$, all strategies show either neutral or positive effects on \ac{auroc5} compared to the unmasked representation, although the differences between selection criteria are relatively small. As more features are masked, the performance curves diverge, with strategy-dependent optimal values before additional feature masking leads to degradation (see \cref{sec:apx-selection-strategies,tab:apx:selection-strategies}).

Across both evaluation domains, selecting features based on positive domain classifier weights provides the most consistent improvement. On \ac{first}, masking the top 256 positive-weight features achieves an \ac{auroc5} of 0.7159, compared to 0.7145 without masking, while on \ac{scanner}, the same strategy reaches 0.7130 compared to approximately 0.7097 for the unmasked representation. The top absolute-weight strategy achieves a comparable maximum performance, reaching 0.7160 on \ac{first} at $k=128$ and 0.7122 on \ac{scanner} at $k=256$, but is less consistent across masking levels.

In contrast, masking features associated with \ac{first} (top negative weights) provides no consistent benefit and leads to substantial performance degradation for large $k$. Interestingly, masking a very large number of positive-weight features ($k=3500$) partially recovers performance, indicating that the relapse signal is not exclusively dependent on the masked domain-specific dimensions.

Overall, we find that positively weighted features capture domain-specific variation and use this criterion for identification in subsequent experiments.

\subsection{Domain and Relapse Features Capture Distinct Information}\label{sec:results-domain-vs-relapse}

Having identified domain-specific features and their impact on relapse prediction, we next analyse task-specific features via highly activating images. We further assess whether domain- and relapse-specific dimensions overlap.

\begin{figure}[tb]
    \centering
    \includegraphics[width=.9\linewidth]{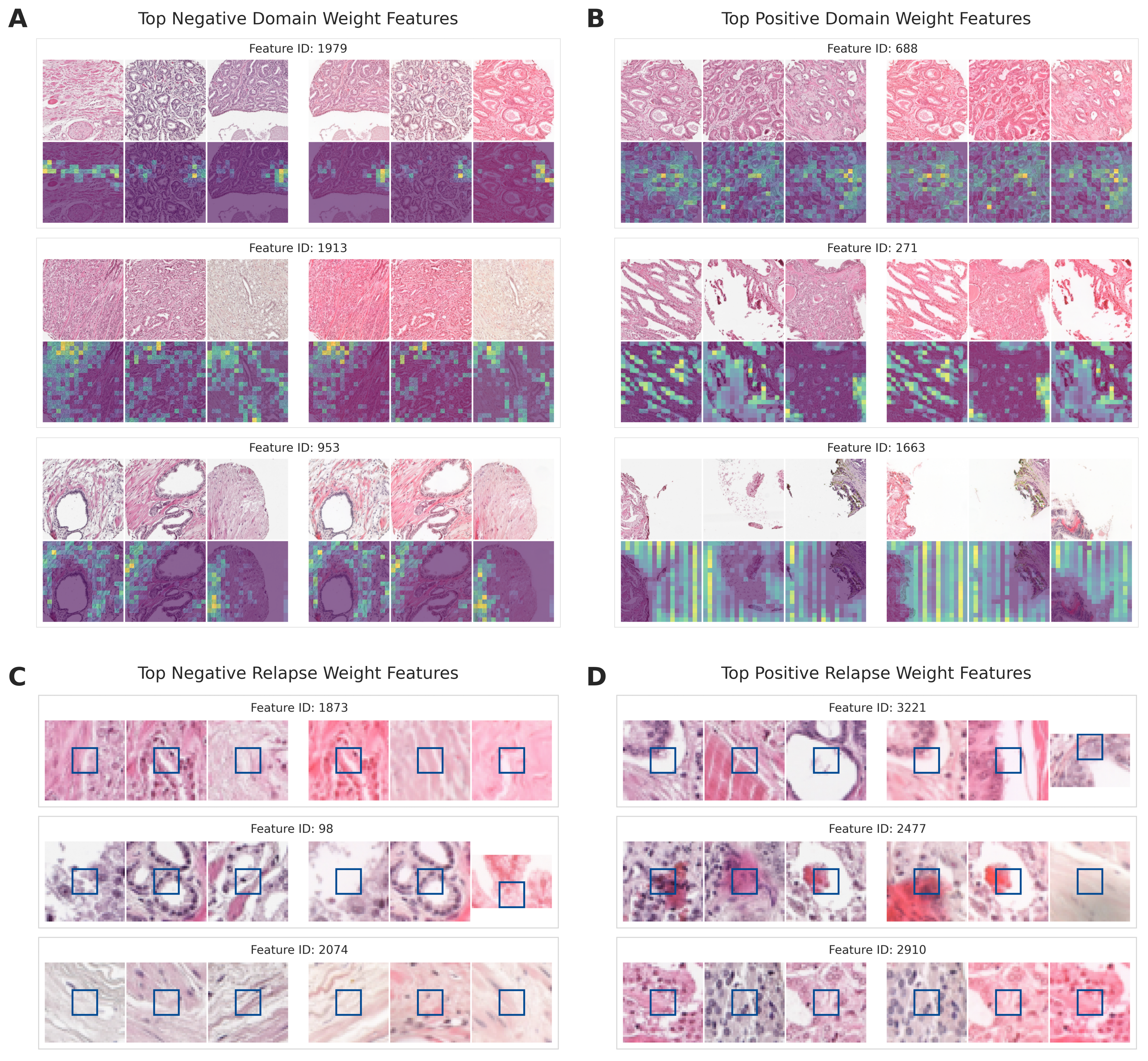}
    \caption{Visualization of highly activating images and patches for domain- and relapse-specific sparse features, respectively. For each ID, examples are chosen from the \ac{first} (left panel) and \ac{scanner} (right panel) subdataset separately. For the top negative (A) and top positive (B) domain-specific features, we show the three most strongly activating \ac{tma} spots together with activation heatmaps. The top negative (C) and top positive (D) relapse-specific features are visualized with three highly activating patches and their surrounding tissue context.}
    \label{fig:explainability}
\end{figure}

\cref{fig:explainability} visualizes highly activating patterns for domain- and relapse-specific sparse features. For domain-specific features (A, B), we show the three most strongly activating \ac{tma} spots from \ac{first} and \ac{scanner}, together with corresponding activation heatmaps. The shown features correspond to the top negative and top positive domain classifier weights, aligning with \ac{first} and \ac{scanner} domain class, respectively. For relapse-specific features (C, D), we show the three most strongly activating patches and their surrounding tissue context, where negative and positive weights correspond to the no-relapse and relapse classes, respectively. In each subplot, the left side shows samples from \ac{first}, while the right side shows the top activating samples from \ac{scanner}.

Domain-specific features (A, B) exhibit highly consistent activation patterns across both domains, often responding to similar or identical tissue regions. This suggests that domain-specific dimensions primarily capture shared visual structures, although some features also activate on background regions, indicating sensitivity to non-biological cues.

Relapse-specific features (C, D) also show strong cross-domain consistency and predominantly focus on coherent tissue structures rather than acquisition artifacts. Overall, both feature groups exhibit substantial overlap across domains, suggesting that sparse features encode largely domain-invariant visual concepts, while domain-specific features more frequently capture dataset-specific shortcuts.

\begin{figure}[tb]
    \centering
    \includegraphics[width=.9\linewidth]{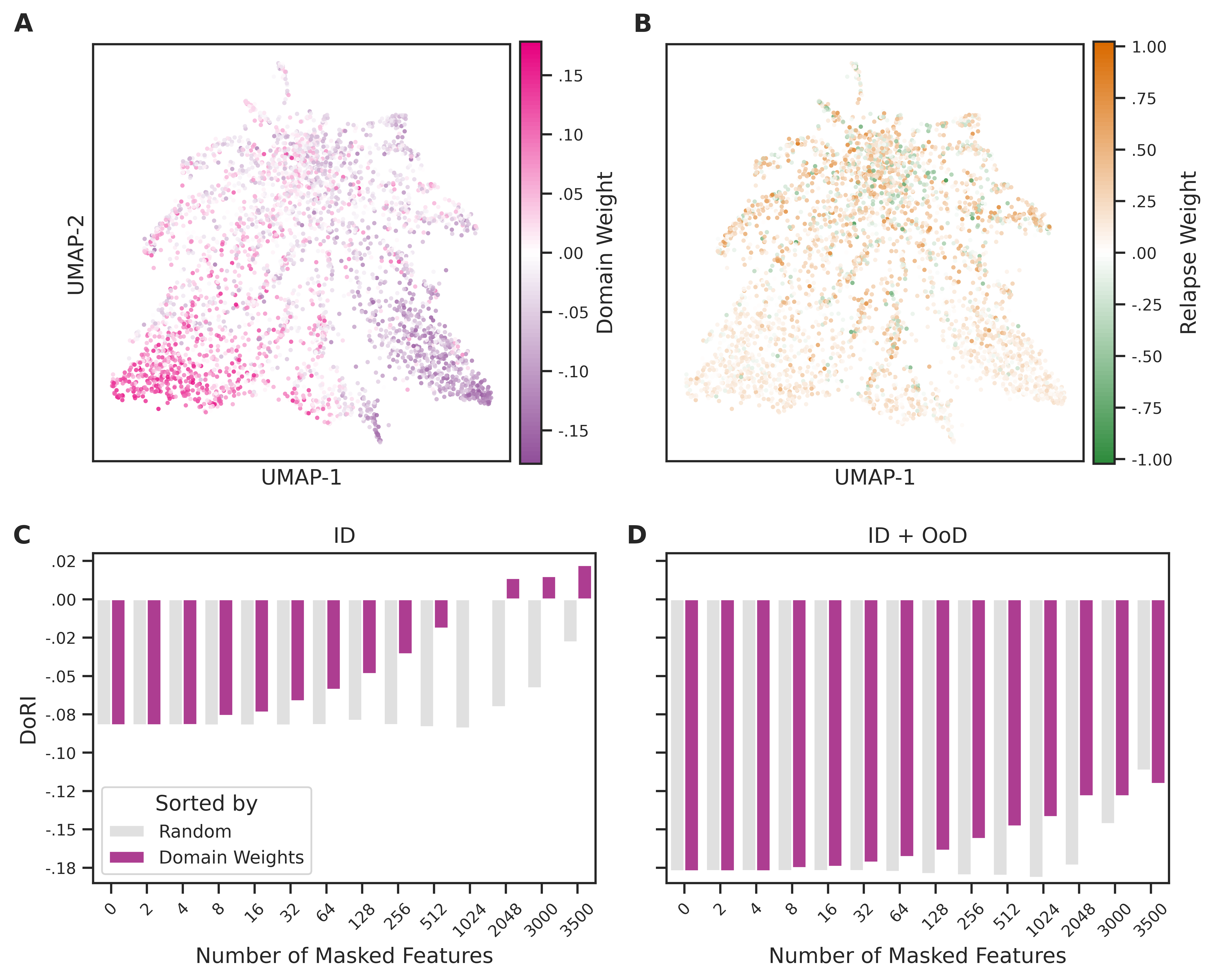}
    \caption{Analysis of the \ac{sae} embedding space. In (A) and (B) the \ac{sae} feature dimension are visualized as activation delta between \ac{first} and \ac{scanner}. The UMAPs are coloured by the feature weights for domain classification (A), and relapse classification~(B). In (C) and (D), we evaluate the Domain Robustness Index (DoRI) for different numbers of masked features. Gray and purple bars represent masked features that are chosen randomly and according to its domain weight, respectively. The \ac{dori} is calculated within the embedding space of in-domain (C) and all subdatasets combined (D).}
    \label{fig:sae-embedding-analysis}
\end{figure}

To further compare both representations, \cref{fig:sae-embedding-analysis} (A and B) projects the pairwise activation delta between \ac{scanner} and \ac{first} samples for all \ac{sae} feature dimensions into a shared embedding space. Note that, for the UMAP visualization, features are treated as observations by performing dimensionality reduction across the image dimension of the feature-by-image activation matrix. Colouring the embedding by domain classifier weights (\cref{fig:sae-embedding-analysis} A) reveals a strong correspondence of the activation difference between scanners, indicating that our identified domain-specific features primarily encode scanner-dependent activation shifts. 
% \begin{figure}[tb]
%     \centering
%     \includegraphics[width=.9\linewidth]{images/umap_delta.png}
%     \caption{Visualization of the \ac{sae} features as activation delta between UKE.scanner and UKE.first. The UMAPs are coloured by the feature weights for domain classification (A), and relapse classification (B).}
%     \label{fig:umap-activation-delta}
% \end{figure}

In contrast, relapse classifier weights (\cref{fig:sae-embedding-analysis} B) are distributed across different regions of the shared embedding space and show little spatial overlap with the domain-specific features. Positive and negative relapse features are concentrated in the embedding region with a low small absolute domain weight. Together, these observations suggest that domain-specific and relapse-specific information are partly disentangled within the learned \ac{sae} representation.

\subsection{Masking Domain-Specific Features Improves Domain Robustness}\label{sec:results-robustness-impact}

We evaluate whether masking domain-specific \ac{sae} features improves robustness under domain shift. We consider both cross-domain prediction performance and changes in representation structure measured by \ac{dori}. This provides a combined view of predictive and embedding-level effects of domain-specific feature masking.

\subsubsection{Cross-Domain Performance.}
We next evaluate whether masking domain-specific features improves generalization across datasets. The change in \ac{auroc5} while an increasing number of domain-specific features is masked is summarized in the appendix (\cref{sec:apx-masking,tab:apx:filter-pos}). Both \ac{id} and \ac{ood} datasets benefit from masking domain-specific features, although the optimal number of masked dimensions differs: \ac{id} performance reaches its maximum after masking a relatively small number of features ($k=256$), whereas \ac{ood} performance continues to improve as additional domain-specific features are masked, reaching its optimum only after substantially more features ($k=3500$) or, in other words, considering only 340 feature dimensions. 

\cref{tab:results-summary} reports the \ac{auroc5} performance of four relapse prediction models: the base model, the unmasked \ac{sae} model, and the two \ac{sae}-based models with different numbers of masked top-$k$ domain-specific features. On \ac{id} data, the unmasked \ac{sae} improves performance by $+0.93$ pp (percentage points), while masking $k=256$ domain-specific features yields the strongest improvement of $+1.18$ pp, followed by $+0.79$ pp when masking $k=3500$ features.

\begin{table}[tb]
    \caption{Cross-domain evaluation of relapse prediction. Results show \ac{auroc5} (mean ± std) for Base, \ac{sae}, and \ac{sae} with masking of domain-specific sparse features ($k=256$, $k=3500$). Highlighted values indicate per-dataset best and second-best performance.\label{tab:results-summary}}
    \centering
    \centering
\setlength{\tabcolsep}{6pt}
\begin{tabular}{lrrrr}
    \toprule
    Dataset & Base & SAE & SAE$^+_{256}$ & SAE$^+_{3500}$ \\
    \midrule
    \ac{first} & 68.20 $\pm$ 0.26 & 69.20 $\pm$ 0.41 & \textbf{69.39} $\pm$ 0.35 & \underline{69.23} $\pm$ 0.08 \\
    \ac{scanner} & 68.58 $\pm$ 0.31 & \underline{69.44} $\pm$ 0.39 & \textbf{69.75} $\pm$ 0.17 & 69.13 $\pm$ 0.16 \\
    \ac{spot} & 65.38 $\pm$ 0.56 & 65.53 $\pm$ 0.76 & \underline{65.83} $\pm$ 0.71 & \textbf{67.74} $\pm$ 0.11 \\
    \ac{thin} & \textbf{60.48} $\pm$ 1.47 & 57.78 $\pm$ 0.48 & 57.80 $\pm$ 0.47 & \underline{58.63} $\pm$ 0.15 \\
    \ac{thick} & 58.17 $\pm$ 0.74 & 58.40 $\pm$ 1.48 & \underline{58.44} $\pm$ 0.69 & \textbf{59.62} $\pm$ 0.17 \\
    \ac{long} & 61.67 $\pm$ 1.09 & 61.79 $\pm$ 1.09 & \underline{62.04} $\pm$ 0.58 & \textbf{62.67} $\pm$ 0.06 \\
    \midrule
    Mean ID & 68.39 $\pm$ 0.34 & \underline{69.32} $\pm$ 0.40 & \textbf{69.57} $\pm$ 0.32 & 69.18 $\pm$ 0.13 \\
    Mean OoD & \underline{61.42} $\pm$ 2.83 & 60.87 $\pm$ 3.31 & 61.03 $\pm$ 3.34 & \textbf{62.16} $\pm$ 3.64 \\
    \bottomrule
\end{tabular}
\end{table}

On \ac{ood} data, the unmasked \ac{sae} decreases performance by $-0.54$ pp, while masking $k=256$ features results in a smaller drop of $-0.39$ pp. In contrast, masking $k=3500$ features improves performance by $+0.75$ pp, suggesting that stronger masking of domain-specific features can improve robustness under unseen domain shifts. The improvement is not uniform across all \ac{ood} domains: \ac{thin} remains challenging, as masking domain-specific features does not improve performance. This indicates that some unseen domain shifts may involve factors not captured by the identified domain-specific sparse features. Notably, the larger standard deviation of the aggregated \ac{ood} performance reflects the varying difficulty across unseen domains.

\subsubsection{Embedding Space Robustness.}
Beyond predictive performance, we evaluate how domain-specific feature masking affects the learned representation using the \ac{dori}, which quantifies the separation of samples according to domain and biological information. \cref{fig:sae-embedding-analysis} (C and D) shows the \ac{dori} as progressively more features are masked for \ac{id} (C) and all subdatasets (D). In both subdataset setups, selecting features according to their domain classifier weights consistently increases the \ac{dori}. In contrast, randomly masking the same number of features produces only minor changes until $k=1024$. 

% \begin{figure}[tb]
%     \centering
%     \includegraphics[width=1\linewidth]{images/dori.pdf}
%     \caption{Visualization of the Domain Robustness Index (DoRI) for different numbers of masked features. Gray and purple bars represent masked features that are chosen randomly and according to its domain weight, respectively. The DoRI is calculated within the embedding space of in-domain subdatasets (A) and all subdatasets combined (B).}
%     \label{fig:dori}
% \end{figure}

\section{Conclusion}\label{sec:conclustion}
We presented \ac{expose}, a framework that uses \ac{sae} as an explainable bottleneck to identifying and masking domain-specific information in \ac{vfm} embeddings for computational pathology. By quantifying domain dependence of sparse features using a linear domain classifier and masking the most predictive dimensions, \ac{expose} enables post-hoc suppression of domain-specific information without modifying or retraining the underlying \ac{vfm}.

Across our multi-domain dataset, selectively masking domain-specific sparse features improved cross-domain relapse prediction compared to using the complete embedding representation. Notably, domain-specific features were identified using only a single pair of acquisition domains, yet their suppression improved robustness across additional unseen domain shifts. This demonstrates that sparse representations contain identifiable domain-specific variation that can be selectively masked to improve cross-domain prediction.

Embedding-space analysis further showed that domain- and relapse-specific information are partially separated within the sparse representation. Domain-specific features aligned with scanner-dependent activation shifts, whereas relapse-specific features occupied different regions of the latent space. Visualization of highly activating images and patches indicated that several sparse dimensions exhibit consistent and explainable activation patterns across domains. Increased \ac{dori} scores after masking further suggest that the improvements result from masking domain-specific variation rather than simply suppressing information.

This study has several limitations. First, we evaluate \ac{expose} using a single \ac{vfm} and a ReLU-based \ac{sae}. Future work should investigate whether similar sparse feature organization emerges across different \acp{vfm} and whether advanced \ac{sae} architectures, such as Gated, Top-$k$, or JumpReLU variants \cite{gatedsae,ksparse,jumprelu}, improve the separation of domain- and task-specific features. Second, domain-specific features are identified using only one pair of acquisition domains. Future studies should evaluate their consistency across additional domain shifts.

Beyond improving cross-domain robustness, \ac{expose} provides an explicit ranking of sparse feature dimensions according to their domain or biological relevance. Future work could further refine this ranking towards a small set of highly informative features that can be rigorously evaluated by expert pathologists, supporting further biological interpretation and the development of more robust computational pathology models.

%\clearpage  % TODO FINAL: This \clearpage needs to be removed from both review and camera-ready versions.

\section*{Acknowledgements}
We would like to thank the IT of the Institute for Medical Systems Bioinformatics and the bAIome Center for Biomedical AI, Sven Heins and Vadim Ustinov.

PF was supported by DFG SFB 1286 SP02. SB received funding from DFG SFB 1713 (C01), SFB 1700 (SP01), and TRR 422 (CP2). JB received funding from DFG SFB 1700 SP01. MZ received funding from DFG SFB 1192 B9. MZ and AW received funding from DFG SFB 1700 A8. AW, JB, and SB were supported by CDL FLIGHT of the University of Hamburg.

\bibliographystyle{splncs04}
\bibliography{main}

\clearpage
\appendix
\section{Supplementary Material} \label{sec:appendix}

\subsection{Additional Dataset Information} \label{sec:apx-data}
\begin{table}[!h]
    \caption{Overview of additional dataset variants derived from the \ac{first} dataset. \ac{first} follows the standard acquisition protocol, while other datasets introduce controlled variations in scanner, staining time, slicing thickness, or sampling location. Differences from the standard protocol are shown in bold. Tissue cores are obtained from two different regions of the prostate. \ac{first} and \ac{scanner} contain the registered images of the same \ac{tma} spots scanned with different scanners. All other subdatasets of tissue core 1 are different slices of the same core.}
    \label{tab:apx:dataset-info}
    \centering
    \small
    \centering
\setlength{\tabcolsep}{4pt}
\begin{tabular}{lllllll}
    \toprule
    Dataset & Images & \parbox{.9cm}{Tissue\\Core} & \parbox{1.4cm}{Slicing\\Thickness} &  \parbox{2.4cm}{Staining Time\\in min (H/E)} & Scanner & Mag. \\
    \midrule
    \ac{first} & 8141 & 1 & 2.5 $\mu$m & 4:00 / 1:20 & Aperio & 40x \\
    \ac{scanner} & 8141 & 1 & 2.5 $\mu$m & 4:00 / 1:20 & \textbf{3DHistech} & \textbf{80x}\\
    \ac{spot} & 8181 & \textbf{2} & 2.5 $\mu$m & 4:00 / 1:20 & Aperio & 40x \\
    \ac{thin} & 1951 & 1 & \textbf{1} $\mu$m & 4:00 / 1:20 & Aperio & 40x \\
    \ac{thick} & 1950 & 1 & \textbf{10} $\mu$m & 4:00 / 1:20 & Aperio & 40x \\
    \ac{long} & 1951 & 1 & 2.5 $\mu$m & \textbf{40:00 / 10:00} & Aperio & 40x \\    
    \bottomrule    
\end{tabular}

\end{table}

\newpage
\subsection{Additional Results on Domain Feature Selection Strategies}\label{sec:apx-selection-strategies}
\begin{table}[!h]
    \caption{Evaluation of the three selection strategies for masking domain-specific sparse features on \ac{first} (a) and \ac{scanner} (b) performance (\ac{auroc5}). Results are shown for validation subdatasets across different numbers of masked features $k$. Highlighted values indicate per-strategy best and second-best performance.}
    \label{tab:apx:selection-strategies}
    \centering
    \noindent\makebox[\textwidth][c]{%
        \begin{minipage}{1\textwidth} % Set this slightly wider than \textwidth
            \centering
            \begin{subtable}[t]{1\textwidth}
                \centering
                \small
                \setlength{\tabcolsep}{10pt}
                \centering
\begin{tabular}{lrrr}
    \toprule
    k & Top Negative & Top Positive & Top Absolute \\
    \midrule
    0 & \underline{71.45} $\pm$ 0.13 & 71.45 $\pm$ 0.13 & 71.45 $\pm$ 0.13 \\
    \midrule
    2 & \underline{71.45} $\pm$ 0.13 & 71.45 $\pm$ 0.14 & 71.46 $\pm$ 0.14 \\
    4 & 71.44 $\pm$ 0.14 & 71.46 $\pm$ 0.14 & 71.47 $\pm$ 0.13 \\
    8 & 71.44 $\pm$ 0.14 & 71.47 $\pm$ 0.13 & 71.48 $\pm$ 0.14 \\
    16 & \textbf{71.46} $\pm$ 0.11 & 71.55 $\pm$ 0.17 & 71.48 $\pm$ 0.14 \\
    32 & \underline{71.45} $\pm$ 0.12 & 71.55 $\pm$ 0.15 & \underline{71.51} $\pm$ 0.13 \\
    128 & 71.38 $\pm$ 0.13 & \underline{71.58} $\pm$ 0.17 & \textbf{71.60} $\pm$ 0.16 \\
    256 & 71.23 $\pm$ 0.14 & \textbf{71.59} $\pm$ 0.20 & 71.44 $\pm$ 0.16 \\
    512 & 71.13 $\pm$ 0.14 & 71.50 $\pm$ 0.18 & 71.29 $\pm$ 0.13 \\
    1024 & 70.97 $\pm$ 0.11 & 71.36 $\pm$ 0.17 & 71.26 $\pm$ 0.09 \\
    2048 & 67.83 $\pm$ 7.14 & 71.01 $\pm$ 0.17 & 70.68 $\pm$ 0.12 \\
    3000 & 70.74 $\pm$ 0.17 & 70.79 $\pm$ 0.10 & 63.65 $\pm$ 13.72 \\
    3500 & 69.72 $\pm$ 0.10 & 71.44 $\pm$ 0.10 & 68.81 $\pm$ 0.05 \\
    \bottomrule
\end{tabular}
                \caption{\ac{first}}
                \label{tab:appx:scana}
            \end{subtable}
                \begin{subtable}[t]{1\textwidth}
                \centering
                \small
                \setlength{\tabcolsep}{10pt}
                \centering
\begin{tabular}{lrrr}
    \toprule
    k & Top Negative & Top Positive & Top Absolute \\
    \midrule
    0 & 70.97 $\pm$ 0.09 & 70.97 $\pm$ 0.09 & 70.97 $\pm$ 0.09 \\
    \midrule
    2 & 70.96 $\pm$ 0.09 & 70.97 $\pm$ 0.09 & 71.01 $\pm$ 0.09 \\
    4 & 70.97 $\pm$ 0.08 & 71.05 $\pm$ 0.09 & 71.01 $\pm$ 0.10 \\
    8 & 70.98 $\pm$ 0.07 & 71.06 $\pm$ 0.08 & 71.10 $\pm$ 0.09 \\
    16 & 70.99 $\pm$ 0.08 & 71.20 $\pm$ 0.12 & 71.11 $\pm$ 0.10 \\
    32 & 70.99 $\pm$ 0.10 & 71.21 $\pm$ 0.12 & 71.13 $\pm$ 0.10 \\
    64 & 71.01 $\pm$ 0.10 & 71.17 $\pm$ 0.13 & 71.15 $\pm$ 0.15 \\
    128 & \textbf{71.04} $\pm$ 0.16 & 71.18 $\pm$ 0.11 & \underline{71.21} $\pm$ 0.15 \\
    256 & 70.81 $\pm$ 0.14 & \textbf{71.30 $\pm$} 0.18 & \textbf{71.22} $\pm$ 0.23 \\
    512 & \underline{71.01} $\pm$ 0.17 & \underline{71.26} $\pm$ 0.24 & 70.96 $\pm$ 0.13 \\
    1024 & 70.66 $\pm$ 0.13 & 70.95 $\pm$ 0.07 & 70.78 $\pm$ 0.07 \\
    2048 & 67.56 $\pm$ 6.80 & 70.47 $\pm$ 0.29 & 70.46 $\pm$ 0.13 \\
    3000 & 70.12 $\pm$ 0.23 & 69.93 $\pm$ 0.15 & 64.08 $\pm$ 13.11 \\
    3500 & 69.25 $\pm$ 0.06 & 70.44 $\pm$ 0.08 & 68.73 $\pm$ 0.04 \\
    \bottomrule
\end{tabular}
                \caption{\ac{scanner}}
                \label{tab:appx:scanb}
            \end{subtable}
        \end{minipage}%
    }    
\end{table}

\newpage
\subsection{Additional Results on Domain Feature Masking}\label{sec:apx-masking}
\begin{table}[!h]
    \caption{Effect of masking domain-specific sparse features on cross-domain relapse prediction performance (\ac{auroc5}). Results are shown for \ac{id} (a) and \ac{ood} (b) evaluation sets across different numbers of masked features $k$. Highlighted values indicate per-dataset best and second-best performance.}
    \label{tab:apx:filter-pos}
    \centering
    \noindent\makebox[\textwidth][c]{%
        \begin{minipage}{1\textwidth} % Set this slightly wider than \textwidth
            \centering
            \begin{subtable}[t]{1\textwidth}
                \centering
                \small
                \setlength{\tabcolsep}{10pt}
                \centering
\begin{tabular}{lrr}
    \toprule
    k & \ac{first} & \ac{scanner}\\
    \midrule
    % test dataset
    0 & 69.20 $\pm$ 0.41 & 69.44 $\pm$ 0.39 \\
    \midrule
    2    & 69.19 $\pm$ 0.41 & 69.42 $\pm$ 0.39 \\
    4    & 69.21 $\pm$ 0.41 & 69.45 $\pm$ 0.40 \\
    8    & 69.20 $\pm$ 0.41 & 69.40 $\pm$ 0.40 \\
    16   & 69.19 $\pm$ 0.41 & 69.37 $\pm$ 0.40 \\
    32   & 69.20 $\pm$ 0.40 & 69.40 $\pm$ 0.40 \\
    64   & 69.19 $\pm$ 0.41 & 69.40 $\pm$ 0.42 \\
    128  & 69.22 $\pm$ 0.43 & 69.39 $\pm$ 0.44 \\
    256  & \textbf{69.39} $\pm$ 0.35 & \textbf{69.75} $\pm$ 0.17 \\
    512  & 69.26 $\pm$ 0.22 & 69.48 $\pm$ 0.47 \\
    1024 & \textbf{69.39} $\pm$ 0.28 & \underline{69.63} $\pm$ 0.22 \\
    2048 & \underline{69.24} $\pm$ 0.38 & 68.79 $\pm$ 0.60 \\
    3000 & 68.93 $\pm$ 0.53 & 68.31 $\pm$ 0.75 \\
    3500 & 69.23 $\pm$ 0.08 & 69.13 $\pm$ 0.16 \\
    \bottomrule
\end{tabular}
                \caption{ID Datasets}
                \label{tab:appx:id}
            \end{subtable}
                \begin{subtable}[t]{1\textwidth}
                \centering
                \small
                \setlength{\tabcolsep}{6pt}
                \centering
 % default is 6pt
\begin{tabular}{lrrrrr}
    \toprule
    k & \ac{spot} & \ac{thin} & \ac{thick} & \ac{long} & Mean OoD\\
    \midrule
        % test dataset
        0 & 65.52 $\pm$ 0.76 & 57.78 $\pm$ 0.48 & 58.40 $\pm$ 1.48 & 61.79 $\pm$ 1.09 & 60.87 $\pm$ 3.31\\
        \midrule
        2    & 65.53 $\pm$ 0.76 & 57.78 $\pm$ 0.48 & 58.40 $\pm$ 1.49 & 61.78 $\pm$ 1.08 & 60.87 $\pm$ 0.95 \\
4    & 65.52 $\pm$ 0.77 & 57.76 $\pm$ 0.48 & 58.42 $\pm$ 1.47 & 61.73 $\pm$ 1.10 & 60.86 $\pm$ 0.95 \\
8    & 65.53 $\pm$ 0.76 & 57.70 $\pm$ 0.49 & 58.37 $\pm$ 1.47 & 61.67 $\pm$ 1.08 & 60.82 $\pm$ 0.95 \\
16   & 65.62 $\pm$ 0.79 & 57.63 $\pm$ 0.50 & 58.28 $\pm$ 1.46 & 61.70 $\pm$ 1.10 & 60.81 $\pm$ 0.96 \\
32   & 65.56 $\pm$ 0.77 & 57.52 $\pm$ 0.49 & 58.24 $\pm$ 1.49 & 61.63 $\pm$ 1.12 & 60.74 $\pm$ 0.97 \\
64   & 65.45 $\pm$ 0.74 & 57.45 $\pm$ 0.46 & 58.16 $\pm$ 1.45 & 61.67 $\pm$ 1.14 & 60.68 $\pm$ 0.95 \\
128  & 65.51 $\pm$ 0.76 & 57.48 $\pm$ 0.45 & 58.13 $\pm$ 1.45 & 61.77 $\pm$ 1.14 & 60.72 $\pm$ 0.95 \\
256  & 65.83 $\pm$ 0.71 & 57.80 $\pm$ 0.47 & 58.44 $\pm$ 0.69 & 62.04 $\pm$ 0.58 & 61.03 $\pm$ 0.61 \\
512  & 65.90 $\pm$ 0.99 & 57.74 $\pm$ 0.72 & 58.42 $\pm$ 0.51 & 61.73 $\pm$ 0.57 & 60.95 $\pm$ 0.70 \\
1024 & 65.30 $\pm$ 0.67 & 56.91 $\pm$ 0.61 & 57.55 $\pm$ 0.75 & 61.07 $\pm$ 0.50 & 60.20 $\pm$ 0.63 \\
2048 & 66.90 $\pm$ 0.57 & \textbf{58.83} $\pm$ 0.83 & 57.76 $\pm$ 1.13 & 61.91 $\pm$ 0.24 & 61.35 $\pm$ 0.69 \\
3000 & \underline{66.96} $\pm$ 0.29 & \underline{58.63} $\pm$ 0.37 & \underline{58.87} $\pm$ 1.07 & \underline{62.20} $\pm$ 0.14 & \underline{61.67} $\pm$ 0.47 \\
3500 & \textbf{67.74} $\pm$ 0.11 & \underline{58.63} $\pm$ 0.15 & \textbf{59.62} $\pm$ 0.17 & \textbf{62.67} $\pm$ 0.06 & \textbf{62.16} $\pm$ 0.12 \\
    \bottomrule
\end{tabular}
                \caption{OoD Datasets}
                \label{tab:appx:ood}
            \end{subtable}
        \end{minipage}%
    }    
\end{table}

\end{document}